\documentclass[conference]{IEEEtran}
\IEEEoverridecommandlockouts

\usepackage{cite}
\usepackage{amsmath,amssymb,amsfonts}
\usepackage{algorithmic}
\usepackage{graphicx}
\graphicspath{{figures/}}
\usepackage{textcomp}
\usepackage{xcolor}
\usepackage{booktabs}
\usepackage{multirow}
\usepackage{url}
\usepackage{microtype}
\usepackage{balance}
\usepackage{tikz}
\usepackage{pgfplots}
\usepgfplotslibrary{groupplots}
\usetikzlibrary{arrows.meta,positioning,fit,calc,backgrounds,shapes.geometric}
\pgfplotsset{compat=1.18}

\definecolor{ieeeblue}{RGB}{0,73,144}
\definecolor{cennblue}{RGB}{42,95,151}
\definecolor{cenngreen}{RGB}{47,120,85}
\definecolor{cennorange}{RGB}{194,109,35}
\definecolor{cennred}{RGB}{176,52,43}
\definecolor{cennlight}{RGB}{245,247,250}
\def\BibTeX{{\rm B\kern-.05em{\sc i\kern-.025em b}\kern-.08em
		T\kern-.1667em\lower.7ex\hbox{E}\kern-.125emX}}

\begin{document}
	
	\title{TinyCeNN-LM: Quality-Gated Conversion of Pretrained Attention with CeNN-Inspired Cellular-Recurrent Layers}
	
	\author{%
		\makebox[\textwidth][c]{%
			\begin{minipage}[t]{0.31\textwidth}\centering
				{\normalsize Kabeh Mohsenzadegan}\par
				{\small\itshape Institute for Smart System Technologies}\par
				{\small University of Klagenfurt}\par
				{\small Klagenfurt, Austria}\par
				{\small kabeh.mohsenzadegan@aau.at}
			\end{minipage}\hfill
			\begin{minipage}[t]{0.31\textwidth}\centering
				{\normalsize Vahid Tavakkoli}\par
				{\small\itshape Institute for Smart System Technologies}\par
				{\small University of Klagenfurt}\par
				{\small Klagenfurt, Austria}\par
				{\small vahid.tavakkoli@aau.at}
			\end{minipage}\hfill
			\begin{minipage}[t]{0.35\textwidth}\centering
				{\normalsize Kyandoghere Kyamakya}\par
				{\small\itshape University of Klagenfurt / Inst. f. Smart Systems Technologies, Austria}\par
				{\small \& Faculte Polytechnique}\par
				{\small Universite de Kinshasa, DR-Congo}\par
				{\small kyandoghere.kyamakya@aau.at}
			\end{minipage}%
		}%
	}
	
	\maketitle
	
	\begin{abstract}
		Replacing attention in a pretrained language model is a compatibility problem: a plausible substitute may alter representations expected by later layers. TinyCeNN-LM introduces a \emph{quality-gated post-training conversion} framework using CeNN-inspired cellular-recurrent layers with bounded local processing, compact recurrent memory, routing, fusion, and accept-or-rollback validation. Three implementations are studied: Integrated Memory, MemoryFusion, and PDelta3-GDN2-CLVR+Local32. Strict PDelta3 conversion accepts a layer only when representation and NLL criteria pass fixed thresholds. On SmolLM2-135M, layers 0--2 are accepted with cumulative $\Delta\mathrm{NLL}=+0.01209$, while layer 3 is rejected despite acceptable NLL because representation fidelity fails. On Qwen3.5-0.8B, full-attention layers 3, 7, and 11 are accepted with final $\Delta\mathrm{NLL}=+0.02073$. Integrated Memory keeps perplexity within $-0.07\%$ to $+0.93\%$ while reducing total cache by up to $6.01\%$. A sampled 200-item downstream sanity check gives $28.5\%$--$32.0\%$ overall accuracy for converted Qwen releases. The results support conservative, quality-gated structural conversion rather than universal attention replacement or speedup.
	\end{abstract}
	
	
	\section{Introduction}
	Softmax attention is highly effective because each token can perform precise content-dependent lookup over earlier representations. Its autoregressive key-value (KV) cache, however, grows with context length. Hardware-aware kernels such as FlashAttention and FlashAttention-2 reduce the I/O cost of exact attention \cite{dao2022flashattention,dao2023flashattention2}, but they do not eliminate the context-dependent state that must be retained during decoding.
	
	Recent sequence models have broadened the design space. Structured state-space models \cite{gu2022s4,gu2023mamba,dao2024mamba2}, retention and recurrent architectures \cite{peng2023rwkv,sun2023retnet,beck2024xlstm}, long convolutions \cite{poli2023hyena}, and hybrid models \cite{arora2024based,lieber2024jamba,botev2024recurrentgemma,ren2024samba,yang2024gateddelta} show that high-capacity sequence modeling need not use full softmax attention at every layer. More recent systems have emphasized learned memory and practical hybrids such as Titans \cite{behrouz2025titans}. Qwen3.5 itself uses a hybrid language backbone combining Gated DeltaNet-style linear recurrence with periodic full-attention anchors \cite{qwen2026qwen35}.
	
	For existing pretrained checkpoints, the central challenge is different from training a new efficient architecture from scratch. Cross-architecture distillation must preserve a function that is already distributed across many layers. RADLADS demonstrates that softmax decoders can be converted to linear-attention decoders with staged distillation \cite{goldstein2025radlads}. KL-guided layer selection further shows that deciding which layers to retain as softmax attention is itself an important optimization problem \cite{li2026klguided}. TinyCeNN-LM addresses the complementary question: \emph{given a proposed replacement at a particular layer, is it faithful enough to commit?}
	
	Our answer is to separate \emph{candidate architecture} from \emph{conversion policy}. Rather than claiming a wholly new universal sequence mixer, TinyCeNN-LM defines a modular cellular-recurrent candidate and asks whether the pretrained network can tolerate it at a specific position. The novelty is strongest in this controlled conversion formulation: each structural change is a hypothesis, evaluated on the current student state and committed only if it satisfies a declared quality contract.
	
\begin{figure}[t]
		\centering
		\includegraphics[width=0.49\textwidth]{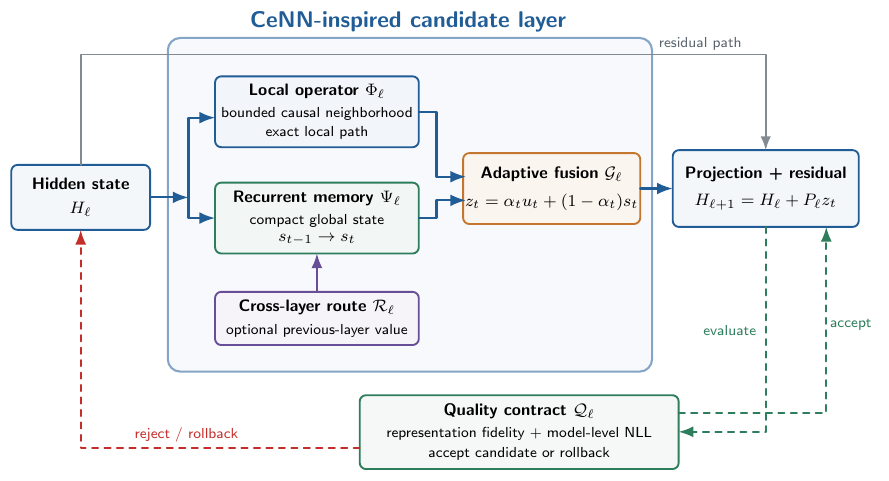}
		\caption{CeNN-inspired conversion layer used throughout TinyCeNN-LM. The local operator $\Phi_{\ell}$ captures bounded neighborhood evidence, the recurrent operator $\Psi_{\ell}$ transports compact long-range state, $\mathcal{R}_{\ell}$ optionally injects cross-layer context, $\mathcal{G}_{\ell}$ fuses the paths, and $\mathcal{Q}_{\ell}$ validates the proposed structural change. The dashed feedback path represents rollback: a candidate that violates the declared quality contract is not committed.}
		\label{fig:abstract}
	\end{figure}

	The contributions are fourfold. First, a five-part CeNN-inspired conversion layer,
	$\mathcal{C}_{\ell}=(\Phi_{\ell},\Psi_{\ell},\mathcal{R}_{\ell},\mathcal{G}_{\ell},\mathcal{Q}_{\ell})$,
	is defined to separate bounded locality, recurrent state, routing, fusion, and validation policy. Second, this abstraction is explicitly connected to classical CeNN principles---local neighborhoods, shared local dynamics, and state evolution---while clarifying the adaptations required for digital token-sequence modeling. Third, a sequential commit/rollback conversion strategy is introduced using a four-metric representation/NLL contract with the same operating thresholds applied across SmolLM2 and Qwen3.5 without per-layer retuning. Finally, cross-model conversion evidence is presented together with the SmolLM2 representation-vs-NLL disagreement, held-out cache/quality measurements for Integrated Memory, and analytical state and parameter accounting for the PDelta3 design.
	Figure~\ref{fig:abstract} summarizes the resulting candidate-layer structure and its quality-gated commit/rollback path.
	
	\section{Related Work}
	\subsection{Efficient Sequence Mixing}
	S4 established structured state-space sequence modeling \cite{gu2022s4}; Hyena explored gated long convolutions \cite{poli2023hyena}; RWKV and RetNet demonstrated recurrent or retention-based inference \cite{peng2023rwkv,sun2023retnet}; and Mamba/Mamba-2 developed selective state spaces and their connection to attention \cite{gu2023mamba,dao2024mamba2}.
	
	Hybrid designs increasingly combine local exactness with compact state. BASED couples linear attention and a small local window \cite{arora2024based}; Jamba, RecurrentGemma, and Samba mix attention with recurrent/state-space blocks \cite{lieber2024jamba,botev2024recurrentgemma,ren2024samba}; xLSTM modernizes recurrent gating \cite{beck2024xlstm}; Gated Delta Networks combine adaptive forgetting with delta-rule updates \cite{yang2024gateddelta}; and Titans explicitly separates short-term attention from learned long-term memory \cite{behrouz2025titans}.
	
	\subsection{Conversion of Pretrained Attention Models}
	Post-training conversion avoids full pretraining cost. RADLADS uses attention/hidden-state alignment, KL distribution matching, and finetuning to distill softmax decoders into linear-attention students \cite{goldstein2025radlads}. KL-guided layer selection estimates which layers should remain softmax under a fixed hybrid budget \cite{li2026klguided}. TinyCeNN-LM does not claim novelty in layer selection alone. Its focus is the \emph{commit decision}: the candidate replacement must satisfy an explicit multi-metric quality contract on the actual student state before it becomes part of the model.
	
	\subsection{From Classical CeNN Locality to Token Sequences}
	Classical cellular neural networks couple a cell state to a bounded neighborhood through shared feedback/feedforward templates \cite{chua1988cnn}. In simplified form,
	\begin{equation}
		\dot{x}_i=-x_i+\sum_{j\in\mathcal{N}(i)}A_{ij}y_j+\sum_{j\in\mathcal{N}(i)}B_{ij}u_j+I_i.
		\label{eq:classicalcenn}
	\end{equation}
	TinyCeNN-LM retains three principles from this view: \emph{(i)} computation is anchored in a bounded neighborhood, \emph{(ii)} information persists through explicit state dynamics, and \emph{(iii)} the same local mechanism is reused across positions. The mapping is not circuit-equivalent. Tokens replace spatial cells; the neighborhood becomes a causal token window; learned projections and input-dependent gates replace fixed analog templates; and a compact recurrent matrix transports information beyond the local window. We therefore use the term \emph{CeNN-inspired} rather than claiming a digital implementation of the original analog differential equation. This locality/state interpretation makes the architectural question explicit: how much of a pretrained attention block can be reconstructed from bounded local interaction plus compact state before functional compatibility is lost?
	
	\section{CeNN-Inspired Layer Design}
	\subsection{Unified Layer Contract}
	Let $H_{\ell,t}\in\mathbb{R}^{d}$ be the hidden representation at layer $\ell$ and token $t$. We define a CeNN-inspired replacement candidate by
	\begin{equation}
		\mathcal{C}_{\ell}=\left(\Phi_{\ell},\Psi_{\ell},\mathcal{R}_{\ell},\mathcal{G}_{\ell},\mathcal{Q}_{\ell}\right),
		\label{eq:abstract}
	\end{equation}
	where $\Phi_{\ell}$ is the bounded local operator, $\Psi_{\ell}$ updates compact recurrent state, $\mathcal{R}_{\ell}$ supplies an optional routed signal, $\mathcal{G}_{\ell}$ fuses local and global evidence, and $\mathcal{Q}_{\ell}$ decides whether the trained candidate is committed.
	
	A generic realization is
	\begin{align}
		u_{\ell,t} &= \Phi_{\ell}(H_{\ell,t-W+1:t}), \\
		r_{\ell,t} &= \mathcal{R}_{\ell}(H_{<\ell,t}), \\
		s_{\ell,t} &= \Psi_{\ell}(s_{\ell,t-1},H_{\ell,t},r_{\ell,t}), \\
		\alpha_{\ell,t} &= \sigma\!\left(g_{\ell}(H_{\ell,t},u_{\ell,t},s_{\ell,t})\right), \\
		z_{\ell,t} &= \alpha_{\ell,t}\odot u_{\ell,t}+(1-\alpha_{\ell,t})\odot s_{\ell,t}, \\
		H_{\ell+1,t} &= H_{\ell,t}+P_{\ell}z_{\ell,t}.
		\label{eq:generic}
	\end{align}
	This representation is deliberately broader than one kernel. It allows the local branch to be exact attention, cellular convolution, or partitioned aggregation; the state branch to be a low-rank recurrent memory or a gated delta state; and routing to be absent or cross-layer.
	
	\subsection{Three Instantiations}
\begin{figure*}[t]
		\centering
		\begin{minipage}[t]{0.47\textwidth}
			\centering
			\includegraphics[width=\linewidth]{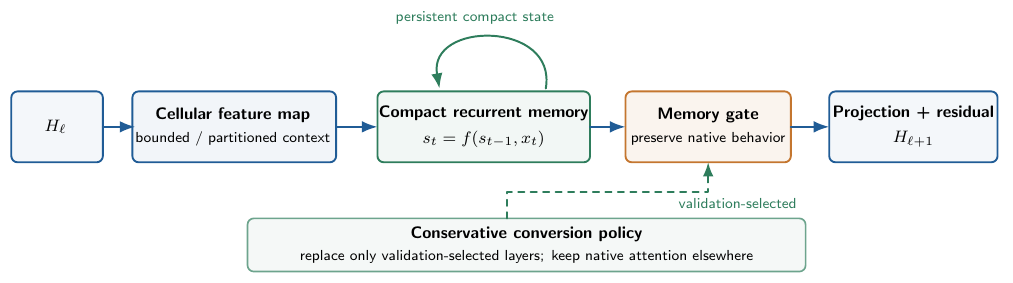}\\[-1mm]
			\textbf{(a) Integrated Memory}
		\end{minipage}\hfill
		\begin{minipage}[t]{0.47\textwidth}
			\centering
			\includegraphics[width=\linewidth]{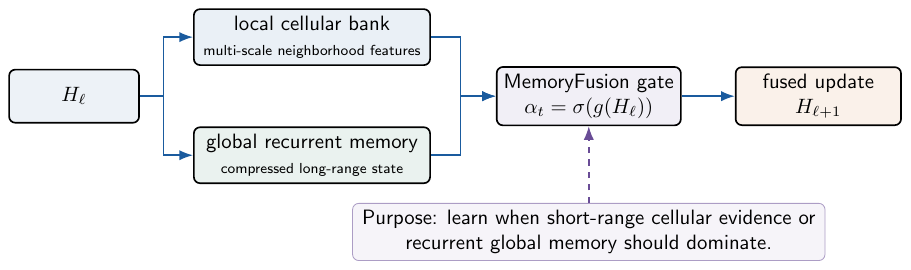}\\[-1mm]
			\textbf{(b) MemoryFusion}
		\end{minipage}
		
		\vspace{1.5mm}
		\begin{minipage}[t]{0.68\textwidth}
			\centering
			\includegraphics[width=\linewidth]{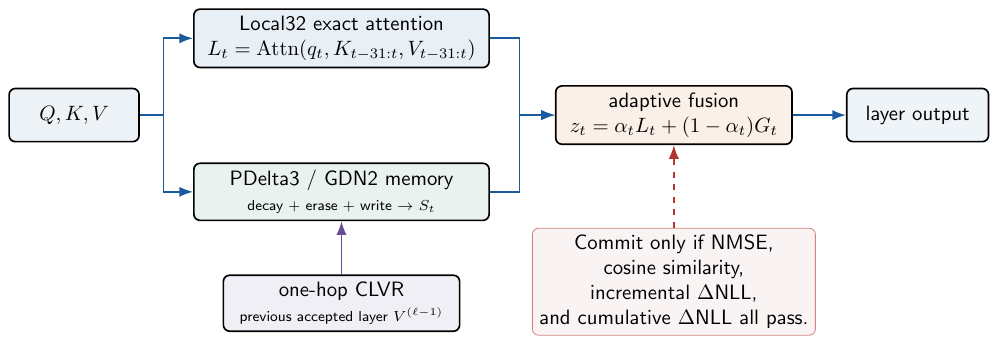}\\[-1mm]
			\textbf{(c) PDelta3-GDN2-CLVR+Local32}
		\end{minipage}
		\caption{Three concrete CeNN-inspired layer instantiations under the abstraction in Fig.~\ref{fig:abstract}. Integrated Memory is the conservative quality/cache design; MemoryFusion is an exploratory local/global fusion design; PDelta3-GDN2-CLVR+Local32 is the strongest strict replacement track in the present experiments.}
		\label{fig:families}
	\end{figure*}

	Figure~\ref{fig:families} shows the three layer families used in the repository.
	
	\textbf{Integrated Memory.} A bounded or partitioned cellular feature map is summarized into compact recurrent memory and merged with the residual stream through a memory gate. Its policy is conservative: only validation-selected layers are converted; native attention is retained elsewhere.
	
	\textbf{MemoryFusion.} A local cellular feature bank and a global recurrent state are computed in parallel. A learned gate decides which path should dominate. This design is useful for studying the local/global allocation itself, although the current repository does not promote its incomplete Qwen strict-gate run as a finished model.
	
	\textbf{PDelta3-GDN2-CLVR+Local32.} This design preserves an exact $W=32$ causal local path and uses GDN2-style recurrent memory globally. For token $t$,
	\begin{equation}
		L_t=\operatorname{softmax}\!\left(\frac{q_tK_{t-W+1:t}^{\top}}{\sqrt{d}}\right)V_{t-W+1:t}.
		\label{eq:local}
	\end{equation}
	The recurrent state uses a gated delta update
	\begin{equation}
		S_t=\left[I-k_t(\beta_t\odot k_t)^\top\right]D_tS_{t-1}
		+k_t(\omega_t\odot \widetilde v_t)^\top,
		\label{eq:gdn}
	\end{equation}
	where $D_t$ is content-dependent decay, $\beta_t$ controls erasure, and $\omega_t$ controls writing. One-hop CLVR modifies the write target for later accepted replacements:
	\begin{equation}
		\widetilde v_t^{(\ell)}=v_t^{(\ell)}+g_t^{route}\odot P_r v_t^{(\ell-1)}.
		\label{eq:clvr}
	\end{equation}
	The two paths are fused head-wise,
	\begin{equation}
		Z_{t,h}=\alpha_{t,h}L_{t,h}+(1-\alpha_{t,h})G_{t,h},\quad
		\alpha_{t,h}=\sigma(q_{t,h}^{\top}w_h+b_h).
		\label{eq:fusion}
	\end{equation}
	
	\section{Quality-Gated Conversion}
	\subsection{Representation and Model-Level Gates}
	Let $A_{\ell}(H_{\ell})$ be the frozen native attention transformation and $R_{\ell}(H_{\ell};\theta_{\ell})$ a candidate CeNN layer evaluated on the \emph{actual current student state}. Representation fidelity is measured by
	\begin{equation}
		\mathrm{NMSE}_{\ell}=
		\frac{\|R_{\ell}(H_{\ell})-A_{\ell}(H_{\ell})\|_2^2}
		{\|A_{\ell}(H_{\ell})\|_2^2+\epsilon},
	\end{equation}
	plus cosine similarity $C_{\ell}$. End-to-end probe degradation is measured by
	\begin{align}
		\Delta\mathrm{NLL}^{inc}_{\ell}&=\mathrm{NLL}_{\ell}-\mathrm{NLL}_{\ell-1},\\
		\Delta\mathrm{NLL}^{cum}_{\ell}&=\mathrm{NLL}_{\ell}-\mathrm{NLL}_{teacher}.
	\end{align}
	The strict quality contract used for the PDelta3 experiments is
	\begin{equation}
		\mathcal{Q}_{\ell}=1 \Longleftrightarrow
		\begin{cases}
			\mathrm{NMSE}_{\ell}\le 0.15,\\
			C_{\ell}\ge 0.94,\\
			\Delta\mathrm{NLL}^{inc}_{\ell}\le 0.015,\\
			\Delta\mathrm{NLL}^{cum}_{\ell}\le 0.05.
		\end{cases}
		\label{eq:gates}
	\end{equation}
	A failure on any metric triggers rollback. The representation gates and NLL gates serve different roles: the former constrain local functional drift, while the latter constrain model-level degradation. Their disagreement is therefore diagnostic. We do not infer from representation mismatch alone that downstream task failure must occur; that stronger causal claim requires a forced-acceptance comparison.
	
	\subsection{Threshold Semantics and Transfer}
	The values in (\ref{eq:gates}) are treated as a \emph{fixed operating point}, not as universal constants. NMSE and cosine constrain magnitude and directional drift of the replaced block; incremental $\Delta$NLL limits damage introduced by the current candidate; cumulative $\Delta$NLL limits error accumulation across an accepted prefix. The same four thresholds are used for the reported SmolLM2 and Qwen3.5 PDelta3 experiments, with no layer-specific threshold relaxation. This cross-model reuse is useful evidence of transfer, but it is not a sensitivity study. A definitive calibration study should vary the four thresholds on a discovery split and report the Pareto frontier on untouched test data.
	
	\subsection{Sequential Commit/Rollback}
	Only one candidate layer is optimized at a time, while previously accepted replacements remain frozen. The training objective combines functional alignment, teacher-logit KL, language-model cross entropy, and a small gate regularizer:
	\begin{equation}
		\mathcal{L}=\lambda_f\mathcal{L}_{align}+\lambda_{KL}\mathcal{L}_{KL}
		+\lambda_{CE}\mathcal{L}_{CE}+\lambda_g\mathcal{L}_{gate}.
	\end{equation}
	The procedure is conceptually simple: train a candidate, evaluate $\mathcal{Q}_{\ell}$, commit and freeze if it passes, otherwise restore the last accepted model. Warm-starting the recurrent core from the previous accepted replacement accelerates later attempts without changing the acceptance rule.
	
	\section{State and Complexity Accounting}
	The system benefit must be separated from the current reference-kernel latency. For one native full-attention layer at batch size one, the autoregressive KV state in $b_s$ bytes per scalar is
	\begin{equation}
		M_{KV}(T)=2Tn_{kv}d_h b_s,
		\label{eq:kvstate}
	\end{equation}
	which grows linearly with decoded context $T$. For an ideal streaming PDelta3-GDN2-CLVR+Local$W$ implementation, the persistent state is
	\begin{equation}
		\begin{aligned}
			M_{CeNN}=b_s\big[&n_{kv}Fd_h+(k-1)(n_h+2n_{kv})d_h\\
			&+2Wn_{kv}d_h\big],
		\end{aligned}
		\label{eq:cennstate}
	\end{equation}
	where the terms correspond to recurrent memory, Q/K/V convolution tails, and a bounded local KV ring buffer. Thus, persistent replacement state is independent of $T$. The dominant sequence-dependent prefill work changes from $\mathcal{O}(T^2n_hd_h)$ for full attention to approximately $\mathcal{O}(TWn_hd_h+Tn_hFd_h)$ for bounded local interaction plus feature-space recurrence; one-token decoding similarly replaces the $T$-dependent attention term by $W$- and $F$-dependent work.
	
	\begin{table}[t]
		\caption{Per-Layer Analytical State for PDelta3+Local32 (FP16, Batch 1)}
		\label{tab:complexity}
		\centering
		\setlength{\tabcolsep}{3.1pt}
		\begin{tabular}{@{}lrrrr@{}}
			\toprule
			Model & Extra params & CeNN state & Native KV@256 & Native KV@2048 \\
			\midrule
			SmolLM2 & 0.153M & 65.6 KiB & 192 KiB & 1.50 MiB \\
			Qwen3.5 & 0.753M & 178 KiB & 512 KiB & 4.00 MiB \\
			\bottomrule
		\end{tabular}
	\end{table}
	The parameter counts in Table~\ref{tab:complexity} are recurrent/routing/fusion parameters added around the retained native Q/K/V/output projections. The state values use $F=96$, $W=32$, $k=4$ with SmolLM2 $(n_h,n_{kv},d_h)=(9,3,64)$ and Qwen3.5 $(8,2,256)$ \cite{benallal2025smollm2,qwen2026qwen35}. They describe the intended streaming architecture, not a measured production PDelta3 cache: the present sequential research wrapper evaluates with \texttt{use\_cache=False}. Measured whole-model cache behavior is therefore reported separately for Integrated Memory in Sec.~\ref{sec:result}; optimized PDelta3 streaming inference remains engineering work.
	
	\section{Experimental Setup}
	\subsection{SmolLM2-135M}
	SmolLM2 provides a compact full-attention testbed \cite{benallal2025smollm2}. The PDelta3 configuration uses feature dimension 96, Local32 attention, recurrent chunk size 32, convolution kernel 4, FP16 state, and an initial local gate of 0.72. Conversion data are streamed from FineWeb-Edu \cite{penedo2024fineweb}. A fixed WikiText-2 probe is held separate from conversion training. The experiment attempts consecutive layers and applies the strict gates in (\ref{eq:gates}).
	
	A separate Integrated Memory V3 experiment trains candidates across contexts $\{256,512,1024\}$ and evaluates held-out contexts $\{256,512,1024,2048\}$. The balanced run uses 128 training, 16 validation, and 32 test documents with deterministic hash-based splitting. Candidate selection uses validation NLL only; the test split is not used for model selection.
	
	\subsection{Qwen3.5-0.8B}
	Qwen3.5-0.8B is a 24-layer hybrid model with periodic full-attention anchors among Gated DeltaNet-style layers \cite{qwen2026qwen35}. The six native full-attention layers in the evaluated text checkpoint are $\{3,7,11,15,19,23\}$. To test whether the same CeNN replacement abstraction transfers beyond a small all-attention model, we target the first three anchors $\{3,7,11\}$ with the same PDelta3-GDN2-CLVR+Local32 structure and the same four quality thresholds. This experiment therefore tests \emph{selective replacement inside an already hybrid backbone}, rather than conversion of every layer.
	
	To add a downstream functional check, we also evaluate standalone Qwen releases with a lightweight sampled FastEval protocol: 50 items each from MMLU-Pro, PIQA, MMMLU-DE, and GPQA-Diamond (200 items total), using CUDA with bfloat16. This is a sampled sanity check rather than an official full-benchmark evaluation. The exported standalone models reload successfully with custom layers $\{3,7,11\}$ for PDelta3-CLVR, $\{3,7\}$ for MemoryFusion, and $\{3,23\}$ for Integrated Memory.
	
	\section{Results}\label{sec:result}
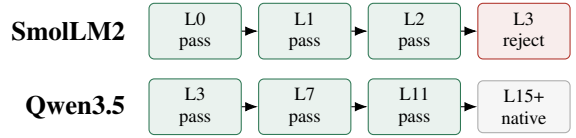
\begin{figure}[t]
		\centering
		\begin{tikzpicture}[font=\scriptsize, x=1cm, y=1cm,
			pass/.style={draw=cenngreen,fill=cenngreen!10,rounded corners=2pt,minimum width=1.22cm,minimum height=.62cm,align=center},
			fail/.style={draw=cennred,fill=cennred!9,rounded corners=2pt,minimum width=1.22cm,minimum height=.62cm,align=center},
			pending/.style={draw=gray!65,fill=gray!7,rounded corners=2pt,minimum width=1.22cm,minimum height=.62cm,align=center},
			arr/.style={-{Latex[length=1.7mm]},thin}]
			\node[anchor=east,font=\bfseries] at (0,1) {SmolLM2};
			\node[pass] (s0) at (0.8,1) {L0\\pass};
			\node[pass] (s1) at (2.25,1) {L1\\pass};
			\node[pass] (s2) at (3.70,1) {L2\\pass};
			\node[fail] (s3) at (5.15,1) {L3\\reject};
			\draw[arr] (s0)--(s1); \draw[arr] (s1)--(s2); \draw[arr] (s2)--(s3);
			
			\node[anchor=east,font=\bfseries] at (0,0) {Qwen3.5};
			\node[pass] (q3) at (0.8,0) {L3\\pass};
			\node[pass] (q7) at (2.25,0) {L7\\pass};
			\node[pass] (q11) at (3.70,0) {L11\\pass};
			\node[pending] (qrest) at (5.15,0) {L15+\\native};
			\draw[arr] (q3)--(q7); \draw[arr] (q7)--(q11); \draw[arr] (q11)--(qrest);
		\end{tikzpicture}
		\caption{Observed quality-gated conversion frontiers. The SmolLM2 experiment exposes a representation-fidelity boundary at layer 3. The Qwen experiment validates the targeted prefix of three full-attention anchors; later anchors are not claimed as converted in this result.}
		\label{fig:frontier}
	\end{figure}

	Table~\ref{tab:crossmodel} summarizes all committed replacements and the first SmolLM2 failure boundary, while Fig.~\ref{fig:frontier} visualizes the corresponding conversion frontiers. On SmolLM2, layers 0--2 pass in the first round. Layer 3 improves over three rounds but remains outside both representation gates. Importantly, its incremental and cumulative NLL are still well below the accepted limits. On Qwen3.5-0.8B, all three targeted full-attention anchors pass, showing that the abstract CeNN replacement can operate inside a native hybrid model rather than only in a small full-attention decoder.
	
	\begin{table*}[t]
		\caption{Strict PDelta3-GDN2-CLVR+Local32 Acceptance Across Two Pretrained Models}
		\label{tab:crossmodel}
		\centering
		\setlength{\tabcolsep}{5.6pt}
		\begin{tabular}{@{}llrrrrrc@{}}
			\toprule
			Model & Layer & NMSE$\downarrow$ & Cosine$\uparrow$ & $\Delta$NLL$_{inc}$ & $\Delta$NLL$_{cum}$ & Local gate & Decision \\
			\midrule
			\multirow{4}{*}{SmolLM2-135M}
			& 0 & 0.0069 & 0.9974 & +0.00239 & +0.00239 & 0.749 & Accept \\
			& 1 & 0.0108 & 0.9960 & +0.00972 & +0.01211 & 0.766 & Accept \\
			& 2 & 0.0856 & 0.9587 & -0.00002 & +0.01209 & 0.766 & Accept \\
			& 3 & \textbf{0.2265} & \textbf{0.8949} & +0.00278 & +0.01487 & 0.441 & Reject \\
			\midrule
			\multirow{3}{*}{Qwen3.5-0.8B}
			& 3  & 0.0621 & 0.9745 & -0.00325 & -0.00325 & 0.721 & Accept \\
			& 7  & 0.0389 & 0.9705 & +0.01327 & +0.01003 & 0.721 & Accept \\
			& 11 & 0.1042 & 0.9411 & +0.01070 & +0.02073 & 0.721 & Accept \\
			\bottomrule
		\end{tabular}
	\end{table*}
	
	Table~\ref{tab:fasteval} summarizes the lightweight 200-item Qwen FastEval run. Its purpose is not to establish a benchmark ranking, but to check whether the released standalone conversions remain usable on tasks outside the NLL acceptance probe. On this sample, the native Qwen3.5-0.8B model scores $30.0\%$ overall in the paired baseline runs. PDelta3-CLVR scores $28.5\%$, MemoryFusion $30.5\%$, and Integrated Memory $32.0\%$. With only 50 examples per benchmark, these differences correspond to a small number of items and should not be interpreted as statistically robust superiority or degradation.
	
	\begin{table}[t]
		\caption{Sampled Qwen3.5-0.8B FastEval Sanity Check (50 Items per Benchmark; 200 Total). These Are Not Official Full-Benchmark Scores.}
		\label{tab:fasteval}
		\centering
		\scriptsize
		\setlength{\tabcolsep}{4.1pt}
		\begin{tabular}{@{}lcrrrrr@{}}
			\toprule
			Model / release & Custom & MMLU & PIQA & MMMLU & GPQA-D & Overall \\
			& layers & Pro &  & DE &  &  \\
			\midrule
			Qwen3.5-0.8B (native) & -- & 12.0 & 48.0 & 34.0 & 26.0 & 30.0 \\
			PDelta3-CLVR+Local32 & 3,7,11 & 14.0 & 48.0 & 30.0 & 22.0 & 28.5 \\
			MemoryFusion & 3,7 & 10.0 & 54.0 & 28.0 & 30.0 & 30.5 \\
			Integrated Memory & 3,23 & 12.0 & 42.0 & 44.0 & 30.0 & 32.0 \\
			\bottomrule
		\end{tabular}
	\end{table}
	
	The same runs expose the current implementation cost. Measured generation throughput is $12.20$ tok/s for PDelta3-CLVR, $8.89$ tok/s for MemoryFusion, and $5.80$ tok/s for Integrated Memory, with peak VRAM of approximately $1.48$, $1.51$, and $1.48$~GiB, respectively. The native Qwen baseline measures $13.37$--$13.78$ tok/s and about $1.47$~GiB in the paired runs. Thus, the sampled task results provide evidence against gross post-conversion functional collapse, but they do not change the systems conclusion: the present reference implementations are not optimized for latency.

	The SmolLM2 layer-3 case is the most diagnostic result. At its final attempt, $\Delta$NLL$_{inc}=+0.00278$ and $\Delta$NLL$_{cum}=+0.01487$, comfortably satisfying the language-model gates. Yet NMSE rises to $0.2265$ and cosine falls to $0.8949$. Under the declared thresholds, an NLL-only policy would commit the layer whereas the full contract rolls it back. What is established here is \emph{metric disagreement}: short-horizon language-model loss can remain acceptable while the replaced block drifts substantially from the function the downstream network was trained to consume. We do not yet claim that the representation gates are proven predictors of downstream task failure. The decisive causal experiment is to force-accept this rejected layer and compare it with rollback on long-context prediction, retrieval, reasoning, code, factuality, instruction following, and subsequent conversion stability.
	
	The Qwen result also suggests that replaceability is not simply ``early layers are easy.'' Layers 3, 7, and 11 are separated by native recurrent/linear layers, yet all three pass the same contract. The relevant unit is therefore the interaction between the candidate layer, its incoming representation, and the surrounding backbone.

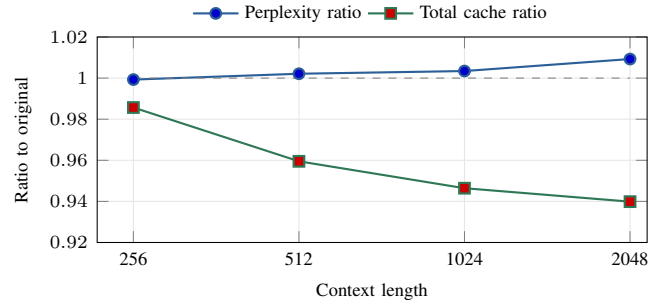
\begin{figure}[t]
		\centering
		\begin{tikzpicture}
			\begin{axis}[
				width=\columnwidth,height=4.3cm,
				xlabel={Context length},ylabel={Ratio to original},
				xmin=220,xmax=2200,ymin=.92,ymax=1.02,
				xmode=log,log basis x=2,
				xtick={256,512,1024,2048},xticklabels={256,512,1024,2048},
				grid=both,major grid style={gray!20},
				tick label style={font=\scriptsize},label style={font=\scriptsize},
				legend style={font=\scriptsize,at={(0.5,1.02)},anchor=south,legend columns=2,draw=none},
				]
				\addplot+[mark=*,thick,cennblue] coordinates {(256,.999279)(512,1.002115)(1024,1.003435)(2048,1.009272)};
				\addlegendentry{Perplexity ratio}
				\addplot+[mark=square*,thick,cenngreen] coordinates {(256,.985677)(512,.959505)(1024,.946419)(2048,.939876)};
				\addlegendentry{Total cache ratio}
				\addplot[dashed,gray] coordinates {(256,1)(2048,1)};
			\end{axis}
		\end{tikzpicture}
		\caption{Integrated Memory V3 preserves test perplexity close to the baseline while the relative total-cache footprint decreases as context grows.}
		\label{fig:cache}
	\end{figure}

	Table~\ref{tab:cache} reports the validation-selected \texttt{partition\_conservative} Integrated Memory candidate on held-out SmolLM2 test documents, and Fig.~\ref{fig:cache} shows the corresponding quality/cache trend across context lengths. Perplexity remains close to the original model while total cache reduction grows with context, reaching $6.01\%$ at 2048 tokens. Because only a small subset of layers is replaced, total-model cache savings are intentionally modest; the experiment measures conservative conversion rather than maximum replacement density.
	
	\begin{table}[t]
		\caption{SmolLM2 Integrated Memory V3 Held-Out Results}
		\label{tab:cache}
		\centering
		\setlength{\tabcolsep}{3.3pt}
		\begin{tabular}{@{}rrrrr@{}}
			\toprule
			Context & PPL/orig. & Cache/orig. & Save & Prefill / Decode \\
			\midrule
			256  & 0.9993 & 0.9857 & 1.43\% & 0.830 / 0.893$\times$ \\
			512  & 1.0021 & 0.9595 & 4.05\% & 0.849 / 0.910$\times$ \\
			1024 & 1.0034 & 0.9464 & 5.36\% & 0.829 / 0.892$\times$ \\
			2048 & 1.0093 & 0.9399 & 6.01\% & 0.830 / 0.891$\times$ \\
			\bottomrule
		\end{tabular}
	\end{table}
	
	The timing columns are intentionally reported as measured: ratios below $1\times$ mean the current PyTorch/reference CeNN implementation is slower than optimized SDPA. At context 2048, prefill is about $0.83\times$ and decode about $0.89\times$ the original speed. No kernel-level speedup is claimed. A fused CUDA/Triton implementation is required before making latency conclusions about the architecture itself.
	
	\section{Discussion}
	Local exact attention, recurrent state, delta updates, gating, and local/global hybrids all have strong precedents. TinyCeNN-LM therefore does not claim that PDelta3 is a fundamentally new sequence mixer. Its main contribution is the \emph{conversion methodology}: architecture is separated from policy, replacements are trained sequentially, representation fidelity and model-level loss are checked jointly, and failed substitutions are rolled back.
	
	KL-guided selection \cite{li2026klguided} identifies promising layers under a hybrid budget, while RADLADS \cite{goldstein2025radlads} studies staged cross-architecture distillation. TinyCeNN-LM addresses the complementary \emph{commit decision} after training a candidate, allowing global layer selection to be combined with a local quality contract.
	
	The results support three main conclusions. First, strict PDelta3 conversion accepts multiple substitutions in both SmolLM2 and Qwen3.5-0.8B. Second, the SmolLM2 frontier shows that acceptable NLL does not always imply local functional fidelity. Third, conservative Integrated Memory reduces whole-model cache usage while preserving perplexity closely. The sampled Qwen FastEval adds a preliminary downstream check: the standalone converted releases score $28.5\%$--$32.0\%$ overall versus $30.0\%$ for the native model in paired runs. Because only 50 items per task are evaluated, this mainly shows that the released conversions remain functional; it is not evidence of benchmark superiority.
	
	The most important next step remains a forced-acceptance experiment at SmolLM2 layer 3, comparing rollback with NLL-only acceptance. Further work should include component ablations, threshold-sensitivity analysis, multi-seed confidence intervals, complete Qwen anchor coverage, and full downstream benchmarks. Current reference kernels remain slower than optimized attention, so the systems claim is structural memory conversion rather than speed superiority.
	
	The TinyCeNN-LM repository provides the implementations, notebooks, checkpoints, and experiment metadata. Canonical runs include the SmolLM2 and Qwen3.5 PDelta3 sequential experiments and SmolLM2 Integrated Memory V3. The accepted Qwen checkpoint is \texttt{vtava/Qwen3.5-0.8B-PDelta3-CLVR-Local32}, and the selected SmolLM2 checkpoint is \texttt{vtava/SmolLM2-135M-CeNN-Partition-V3} \cite{tinycennrepo}.
	
	\section{Conclusion}
	TinyCeNN-LM reframes attention replacement as controlled post-training architecture conversion. Its CeNN-inspired abstraction retains the classical emphasis on local neighborhoods and explicit state while adapting those principles to causal token sequences through learned digital operators, recurrent memory, routing, and fusion. The main methodological contribution is the separation of candidate architecture from a strict commit/rollback contract. Cross-model experiments show successful selective conversion on SmolLM2 and Qwen3.5 and expose a SmolLM2 frontier where NLL-based and representation-based decisions disagree. Integrated Memory further demonstrates a modest but real context-dependent cache/quality tradeoff. A sampled 200-item Qwen downstream sanity check further indicates that the standalone converted releases remain functional beyond the NLL probe, but it is too small to support a benchmark-ranking claim. The present evidence supports conservative structural conversion; it does not yet prove downstream causal benefit or runtime speedup. Forced-acceptance validation, matched ablations, full Qwen anchor coverage, threshold calibration, multi-seed statistics, and optimized streaming kernels are the steps needed to turn this framework into a stronger general claim about reliable Transformer-to-cellular/recurrent conversion.

\end{document}